\documentclass[preprint,11pt]{elsarticle}

\usepackage{graphicx, paralist, amssymb, hyperref, caption, array, verbatim, amsthm}

\usepackage{longtable}
\usepackage{float}
\usepackage[left=2.5cm,top=2.5cm,right=2.5cm,bottom=2.5cm]{geometry}

\usepackage{multicol}
\usepackage{bookmark}

\journal{Zeitschrift f\"{u}r Medizinische Physik}

\begin{document}

\begin{frontmatter}

\title{An overview of deep learning in medical imaging \\focusing on MRI}

\author[mmiv,hvl]{Alexander Selvikv\aa{}g Lundervold\corref{cor1}}
\ead{allu@hvl.no}
\author[mmiv,uib,hvl2]{Arvid Lundervold}
\ead{mfyal@uib.no}

\address[mmiv]{Mohn Medical Imaging and Visualization Centre (MMIV), Haukeland University Hospital, Norway}
\address[hvl]{Department of Computing, Mathematics and Physics, Western Norway University of Applied Sciences, Norway}
\address[uib]{Neuroinformatics and Image Analysis Laboratory, Department of Biomedicine, University of Bergen, Norway}
\address[hvl2]{Department of Health and Functioning, Western Norway University of Applied Sciences, Norway}

\cortext[cor1]{Corresponding author}

\begin{abstract}
What has happened in machine learning lately, and what does it mean for the future of medical image analysis? 
Machine learning has witnessed a tremendous amount of attention over the last few years. The current boom started around 2009 when so-called deep artificial neural networks began outperforming other established models on a number of important benchmarks. Deep neural networks are now the state-of-the-art machine learning models across a variety of areas, from image analysis to natural language processing, and widely deployed in academia and industry. These developments have a huge potential for medical imaging technology, medical data analysis, medical diagnostics and healthcare in general, slowly being realized. 
We provide a short overview of recent advances and some associated challenges in machine learning applied to medical image processing and image analysis. 
As this has become a very broad and fast expanding field we will not survey the entire landscape of applications, but put particular focus on deep learning in MRI. 

Our aim is threefold:
(i) give a brief introduction to deep learning with pointers to core references; (ii) indicate how deep learning has been applied to the entire MRI processing chain, from acquisition to image retrieval, from segmentation to disease prediction;
(iii)  provide a starting point for people interested in experimenting and perhaps contributing to the field of machine learning for medical imaging by pointing out good educational resources, state-of-the-art open-source code, and interesting sources of data and problems related medical imaging.

\end{abstract}


\end{frontmatter}

\section{Introduction}
\label{S:1}

Machine learning has seen some dramatic developments recently, leading to a lot of interest from industry, academia and popular culture. These are driven by breakthroughs in \emph{artificial neural networks}, often termed \emph{deep learning}, a set of techniques and algorithms that enable computers to discover complicated patterns in large data sets. Feeding the breakthroughs is the increased access to data (``big data"), user-friendly software frameworks, and an explosion of the available compute power, enabling the use of neural networks that are deeper than ever before. These models nowadays form the state-of-the-art approach to a wide variety of problems in computer vision, language modeling and robotics.

Deep learning rose to its prominent position in computer vision when neural networks started outperforming other methods on several high-profile image analysis benchmarks. Most famously on the ImageNet Large-Scale Visual Recognition Challenge (ILSVRC)\footnote{Colloquially known as the ImageNet challenge} in 2012 \cite{NIPS2012_4824} when a deep learning model (a \emph{convolutional neural network}) halved the second best error rate on the image classification task. Enabling computers to recognize objects in natural images was until recently thought to be a very difficult task, but by now convolutional neural networks have surpassed even human performance on the ILSVRC, and reached a level where the ILSVRC classification task is essentially solved (i.e. with error rate close to the Bayes rate). The last ILSVRC competition was held in 2017, and computer vision research has moved on to other more difficult benchmark challenges. For example the Common Objects in Context Challenge (COCO) \cite{lin2014microsoft}. 

Deep learning techniques have become the de facto standard for a wide variety of computer vision problems. They are, however, not limited to image processing and analysis but are outperforming other approaches in areas like natural language processing \cite{peters2018deep, howard2018universal, radford2018improving}, speech recognition and synthesis \cite{Xiong2018,Oord2016}\footnote{Try it out here: \url{https://deepmind.com/blog/wavenet-generative-model-raw-audio}}, and in the analysis of unstructured, tabular-type data using \textit{entity embeddings} \cite{guo2016entity, de2015artificial}.\footnote{As a perhaps unsurprising side-note, these modern deep learning methods have also entered the field of physics. Among other things, they are tasked with learning physics from raw data when no good mathematical models are available. For example in the analysis of gravitational waves where deep learning has been used for classification \cite{george2018deep}, anomaly detection \cite{george2018classification} and denoising \cite{shen2017denoising}, using methods that are highly transferable across domains (think EEG and fMRI). They are also part of mathematical model and machine learning hybrids \cite{raissi2018hidden, Karpatne2017}, formed to reduce computational costs by having the mathematical model train a machine learning model to perform its job, or to improve the fit with observations in settings where the mathematical model can't incorporate all details (think noise).}  

The sudden progress and wide scope of deep learning, and the resulting surge of attention and multi-billion dollar investment, has led to a virtuous cycle of improvements and investments in the entire field of machine learning. It is now one of the hottest areas of study world-wide \cite{gartner2018}, and people with competence in machine learning are highly sought-after by both industry and academia\footnote{See e.g. \url{https://economicgraph.linkedin.com/research/LinkedIns-2017-US-Emerging-Jobs-Report} for a study focused on the US job market}. 

Healthcare providers generate and capture enormous amounts of data containing extremely valuable signals and information, at a pace far surpassing what ``traditional" methods of analysis can process. Machine learning therefore quickly enters the picture, as it is one of the best ways to integrate, analyze and make predictions based on large, heterogeneous data sets (cf. health informatics \cite{Ravi2017}). Healthcare applications of deep learning range from one-dimensional biosignal analysis \cite{Ganapathy2018} and the prediction of medical events, e.g. seizures \cite{Kuhlmann2018} and cardiac arrests \cite{Kwon2018}, to computer-aided detection \cite{Shin2016} and diagnosis \cite{Kermany2018} supporting clinical decision making and survival analysis \cite{Katzman2018}, to drug discovery \cite{Jimenez2018} and as an aid in therapy selection and pharmacogenomics \cite{Kalinin2018}, to increased operational efficiency \cite{Jiang2018a}, stratified care delivery \cite{Vranas2017}, and analysis of electronic health records \cite{rajkomar2018scalable, Shickel2017}.

The use of machine learning in general and deep learning in particular within healthcare is still in its infancy, but there are several strong initiatives across academia, and multiple large companies are pursuing healthcare projects based on machine learning. Not only medical technology companies, but also for example Google Brain \cite{gulshan2016development,46425, poplin2018aus}\footnote{\url{https://ai.google/research/teams/brain/healthcare-biosciences}}, DeepMind \cite{de2018clinically}\footnote{\url{https://deepmind.com/applied/deepmind-health/}}, Microsoft \cite{qin2018autofocus, kamnitsas2017unsupervised}\footnote{\url{https://www.microsoft.com/en-us/research/research-area/medical-health-genomics}}  and IBM \cite{xiao2018opportunities}\footnote{\url{https://www.research.ibm.com/healthcare-and-life-sciences}}. There is also a plethora of small and medium-sized businesses in the field\footnote{Aidoc, Arterys, Ayasdi, Babylon Healthcare Services, BenevolentAI, Enlitic, EnvoiAI, H2O, IDx, MaxQ AI, Mirada Medical, Viz.ai, Zebra Medical Vision, and many more.}.

\section{Machine learning, artificial neural networks, deep learning}
\label{S:2}
In machine learning one develops and studies methods that give computers the ability to solve problems by learning from experiences. The goal is to create mathematical models that can be \textit{trained} to produce useful outputs when fed input data. Machine learning models are provided experiences in the form of \textit{training data}, and are tuned to produce accurate predictions for the training data by an optimization algorithm. The main goal of the models are to be able to \textit{generalize} their learned expertise, and deliver correct predictions for new, unseen data. A model's generalization ability is typically estimated during training using a separate data set, the validation set, and used as feedback for further tuning of the model. After several iterations of training and tuning, the final model is evaluated on a test set, used to simulate how the model will perform when faced with new, unseen data.

There are several kinds of machine learning, loosely categorized according to how the models utilize its input data during training. In \textit{reinforcement learning} one constructs \textit{agents} that learn from their environments through trial and error while optimizing some objective function. A famous recent application of reinforcement learning is AlphaGo and AlphaZero \cite{silver2017mastering}, the Go-playing machine learning systems developed by DeepMind. In \textit{unsupervised learning} the computer is tasked with uncovering patterns in the data without our guidance. Clustering is a prime example. Most of today's machine learning systems belong to the class of \textit{supervised learning}. Here, the computer is given a set of already labeled or annotated data, and asked to produce correct labels on new, previously unseen data sets based on the rules discovered in the labeled data set. From a set of input-output examples, the whole model is trained to perform specific data-processing tasks. Image annotation using human-labeled data, e.g. classifying skin lesions according to malignancy \cite{Esteva2017} or discovering cardiovascular risk factors from retinal fundus photographs \cite{poplin2018prediction}, are two examples of the multitude of medical imaging related problems attacked using supervised learning. 

Machine learning has a long history and is split into many sub-fields, of which deep learning is the one currently receiving the bulk of attention.  

There are many excellent, openly available overviews and surveys of deep learning. For short general introductions to deep learning, see \cite{lecun2015deep, Hinton2018}. For an in-depth coverage, consult the freely available book \cite{Goodfellow-et-al-2016}\footnote{\url{https://www.deeplearningbook.org/}}. For a broad overview of deep learning applied to medical imaging, see \cite{Litjens2017}. We will only mention some bare essentials of the field, hoping that these will serve as useful pointers to the areas that are currently the most influential in medical imaging.

\subsection{Artificial neural networks}
Artificial neural networks (ANNs) is one of the most famous machine learning models, introduced already in the 1950s, and actively studied since \cite[Chapter 1.2]{Goodfellow-et-al-2016}.\footnote{The loose connection between artificial neural networks and neural networks in the brain is often mentioned, but quite over-blown considering the complexity of biological neural networks. However, there is some interesting recent work connecting neuroscience and artificial neural networks, indicating an increase in the cross-fertilization between the two fields \cite{Marblestone2016, Hassabis2017, banino2018vector}.} 

Roughly, a neural network consists of a number of connected computational units, called \textit{neurons}, arranged in layers. There's an input layer where data enters the network, followed by one or more \textit{hidden layers} transforming the data as it flows through, before ending at an output layer that produces the neural network's predictions. The network is trained to output useful predictions by identifying patterns in a set of labeled training data, fed through the network while the outputs are compared with the actual labels by an \textit{objective function}. During training the network's parameters--the strength of each neuron--is tuned until the patterns identified by the network result in good predictions for the training data. Once the patterns are learned, the network can be used to make predictions on new, unseen data, i.e. generalize to new data.

It has long been known that ANNs are very flexible, able to model and solve complicated problems, but also that they are difficult and very computationally expensive to train.\footnote{According to the famous \textit{universal approximation theorem} for artificial neural networks \cite{cybenko1989approximation, hornik1989multilayer, leshno1993multilayer, sonoda2017neural}, ANNs are mathematically able to approximate any continuous function defined on compact subspaces of $\mathbb{R}^n$, using finitely many neurons. There are some restrictions on the activation functions, but these can be relaxed (allowing for ReLUs for example) by restricting the function space. This is an existence theorem and successfully \textit{training} a neural network to approximate a given function is another matter entirely. However, the theorem does suggests that neural networks are reasonable to study and develop further, at least as an engineering endeavour aimed at realizing their theoretical powers.} This has lowered their practical utility and led people to, until recently, focus on other machine learning models. But by now, artificial neural networks form one of the dominant methods in machine learning, and the most intensively studied. This change is thanks to the growth of \textit{big data}, powerful processors for parallel computations (in particular, GPUs), some important tweaks to the algorithms used to construct and train the networks, and the development of easy-to-use software frameworks. The surge of interest in ANNs leads to an incredible pace of developments, which also drives other parts of machine learning with it.

The freely available books \cite{Goodfellow-et-al-2016, nielsen2015neural} are two of the many excellent sources to learn more about artificial neural networks. We'll only give a brief indication of how they are constructed and trained. The basic form of artificial neural networks\footnote{These are basic when compared to for example \textit{recurrent neural networks}, whose architectures are more involved}, the \textit{feedforward neural networks}, are parametrized mathematical functions $y = f(\mathbf{x}; \theta)$ that maps an input $\mathbf{x}$ to an output $\mathbf{y}$ by feeding it through a number of nonlinear transformations:
$f(\mathbf{x}) = (f_n \circ \cdots \circ f_1)(\mathbf{x}).$
Here each component $f_k$, called a network \textit{layer}, consists of a simple linear transformation of the previous component's output, followed by a nonlinear function: $f_k = \sigma_k(\theta_k^T f_{k-1})$. The nonlinear functions $\sigma_k$ are typically sigmoid functions or ReLUs, as discussed below, and the $\theta_k$ are matrices of numbers, called the model's \textit{weights}. During the training phase, the network is fed training data and tasked with making predictions at the output layer that match the known labels, each component of the network producing an expedient representation of its input. It has to learn how to best utilize the intermediate representations to form a complex hierarchical representation of the data, ending in correct predictions at the output layer. Training a neural network means changing its weights to optimize the outputs of the network. This is done using an optimization algorithm, called \textit{gradient descent}, on a function measuring the correctness of the outputs, called a \textit{cost function} or \textit{loss function}. The basic ideas behind training neural networks are simple: as training data is fed through the network, compute the gradient of the loss function with respect to every weight using the chain rule, and reduce the loss by changing these weights using gradient descent. But one quickly meets huge computational challenges when faced with complicated networks with thousands or millions of parameters and an exponential number of paths between the nodes and the network output. The techniques designed to overcome these challenges gets quite complicated. See \cite[Chapter 8]{Goodfellow-et-al-2016} and \cite[Chapter 3 and 4]{aggarwal2018neural} for detailed descriptions of the techniques and practical issues involved in training neural networks.

\begin{figure}[h!]
\begin{tabular}{ll}
\hspace{-15mm} & \includegraphics[width=1.10\linewidth]{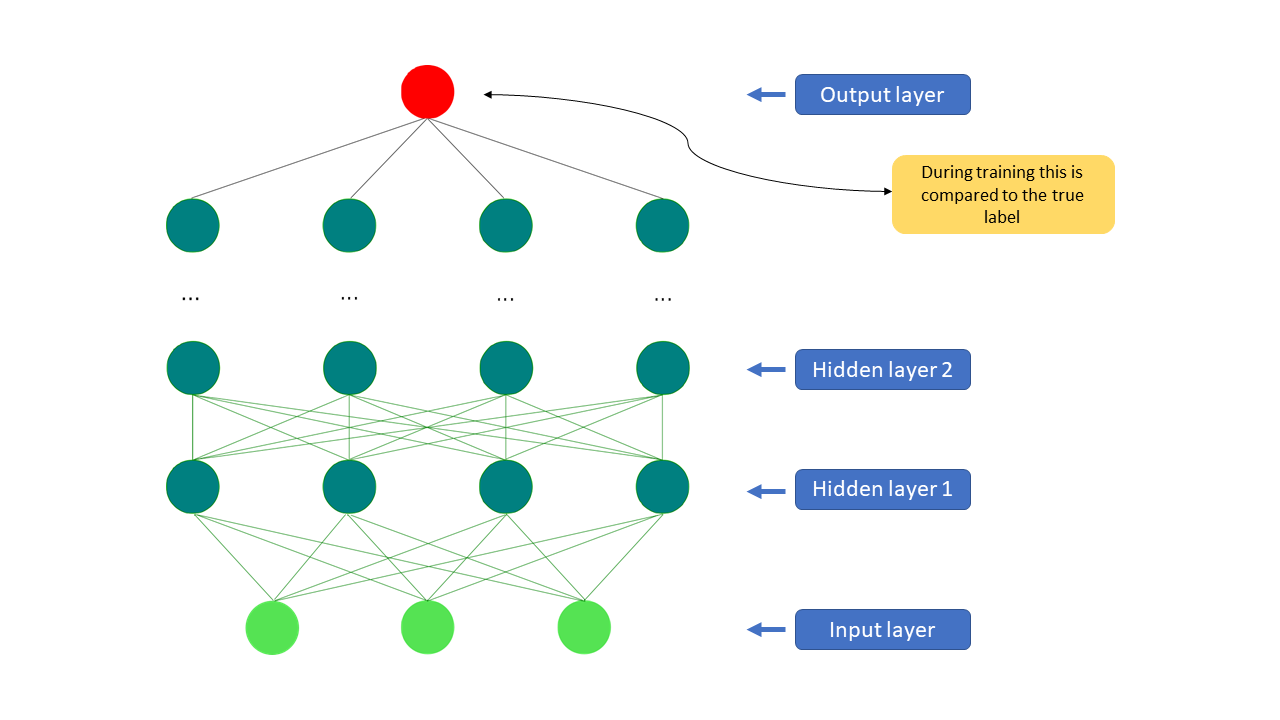}
\end{tabular}
\vspace{-5mm}
\caption{Artificial neural networks are built from simple linear functions followed by nonlinearities. One of the simplest class of neural network is the \textit{multilayer perceptron}, or \textit{feedforward neural network}, originating from the work of Rosenblatt in the 1950s \protect\cite{rosenblatt1958perceptron}. It's based on simple computational units, called \textit{neurons}, organized in \textit{layers}. Writing $i$ for the $i$-th layer and $j$ for the $j$-th unit of that layer, the output of the $j$-th unit at the $i$-th layer is $z_j^{(i)} = \left(\theta_j^{(i)}\right)^T x$. Here $x$ consists of the outputs from the previous layer after they are fed through a simple nonlinear function called an \textit{activation function}, typically a sigmoid function $\sigma(z) = 1/(1+e^{-z})$ or a rectified linear unit $\mbox{ReLU}(z) = \max(0, z)$ or small variations thereof. Each layer therefore computes a weighted sum of the all the outputs from the neurons in the previous layers, followed by a nonlinearity. These are called the \textit{layer activations}. Each layer activation is fed to the next layer in the network, which performs the same calculation, until you reach the \textit{output layer}, where the network's predictions are produced. In the end, you obtain a hierarchical representation of the input data, where the earlier features tend to be very general, getting increasingly specific towards the output. By feeding the network training data, propagated through the layers, the network is \textit{trained} to perform useful tasks. A training data point (or, typically, a small batch of training points) is fed to the network, the outputs and local derivatives at each node are recorded, and the difference between the output prediction and the true label is measured by an \textit{objective function}, such as 
\textit{mean absolute error} (L1),
\textit{mean squared error} (L2),
\textit{cross-entropy loss}, % $L(z, y) = -(y\log(z) + (1-y)\log(1-z))$. 
or \textit{Dice loss}, depending on the application.
The derivative of the objective function with respect to the output is calculated, and used as a feedback signal. The discrepancy is propagated backwards through the network and all the weights are updated to reduce the error. This is achieved using \textit{backward propagation} \protect\cite{linnainmaa1970representation, werbos1974beyond, rumelhart1986learning}, which calculates the gradient of the objective function with respect to the weights in each node using the chain rule together with dynamic programming, and \textit{gradient descent} \protect\cite{cauchy1847methode}, an optimization algorithm tasked with improving the weights.}
\label{fig:neuralnet}
\end{figure}

Artificial neural networks are often depicted as a network of nodes, as in Figure \ref{fig:neuralnet}.\footnote{As we shall see, modern architectures are often significantly more complicated than captured by the illustration and equations above, with connections between non-consecutive layers, input fed in also at later layers, multiple outputs, and much more.}

\subsection{Deep learning} Traditionally, machine learning models are trained to perform useful tasks based on manually designed features extracted from the raw data, or features learned by other simple machine learning models. In deep learning, the computers learn useful representations and features automatically, directly from the raw data, bypassing this manual and difficult step. By far the most common models in deep learning are various variants of artificial neural networks, but there are others. The main common characteristic of deep learning methods is their focus on \textit{feature learning}: automatically learning representations of data. This is the primary difference between deep learning approaches and more ``classical" machine learning. Discovering features and performing a task is merged into one problem, and therefore both improved during the same training process. See \cite{lecun2015deep} and \cite{Goodfellow-et-al-2016} for general overviews of the field. 

In medical imaging the interest in deep learning is mostly triggered by \textit{convolutional neural networks} (CNNs) \cite{lecun1998gradient}\footnote{Interestingly, CNNs was applied in medical image analysis already in the early 90s, e.g. \cite{Lo1993}, but with limited success.}, a powerful way to learn useful representations of images and other structured data. Before it became possible to use CNNs efficiently, these features typically had to be engineered by hand, or created by less powerful machine learning models. Once it became possible to use features learned directly from the data, many of the handcrafted image features were typically left by the wayside as they turned out to be almost worthless compared to feature detectors found by CNNs.\footnote{However, combining hand-engineered features with CNN features is a very reasonable approach when low amounts of training data makes it difficult to learn good features automatically} There are some strong preferences embedded in CNNs based on how they are constructed, which helps us understand why they are so powerful. Let us therefore take a look at the building blocks of CNNs.

\begin{figure}[h!]
\centering\includegraphics[width=1.0\linewidth]{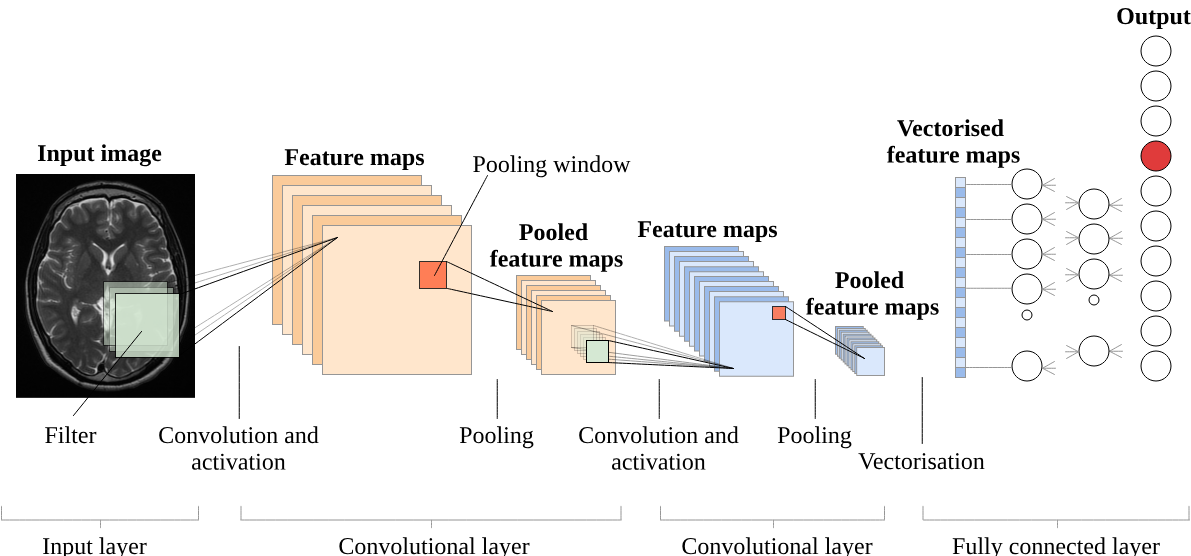}
\caption{{\em Building blocks of a typical CNN.} A slight modification of a figure in \protect\cite{murray2018}, courtesy of the author. }
\end{figure}

\subsection{Building blocks of convolutional neural networks} When applying neural networks to images one can in principle use the simple feedforward neural networks discussed above. However, having connections from all nodes of one layer to all nodes in the next is extremely inefficient. A careful pruning of the connections based on domain knowledge, i.e. the structure of images, leads to much better performance. A CNN is a particular kind of artificial neural network aimed at preserving spatial relationships in the data, with very few connections between the layers. The input to a CNN is arranged in a grid structure and then fed through layers that preserve these relationships, each layer operation operating on a small region of the previous layer (Fig. 2). CNNs are able to form highly efficient representation of the input data\footnote{It's interesting to compare this with the biological vision systems and their \textit{receptive fields} of variable size (volumes in visual space) of neurons at different hierarchical levels}, well-suited for image-oriented tasks. A CNN has multiple layers of \textit{convolutions} and \textit{activations}, often interspersed with \textit{pooling} layers, and is trained using backpropagation and gradient descent as for standard artificial neural networks. See Section 2.1. In addition, CNNs typically have fully-connected layers at the end, which compute the final outputs.\footnote{Lately, so-called \textit{fully-convolution} CNNs have become popular, in which average pooling across the whole input after the final activation layer replaces the fully-connected layers, significantly reducing the total number of weights in the network.}

\begin{enumerate}[i)]
    \item \textbf{Convolutional layers}: In the convolutional layers the activations from the previous layers are convolved with a set of small parameterized \textit{filters}, frequently of size $3 \times 3$, collected in a tensor $W^{(j, i)}$, where $j$ is the filter number and $i$ is the layer number. By having each filter share the exact same weights across the whole input domain, i.e. translational equivariance at each layer, one achieves a drastic reduction in the number of weights that need to be learned. The motivation for this weight-sharing is that features appearing in one part of the image likely also appear in other parts. If you have a filter capable of detecting horizontal lines, say, then it can be used to detect them wherever they appear. Applying all the convolutional filters at all locations of the input to a convolutional layer produces a tensor of \textit{feature maps}. 
    \item \textbf{Activation layer}: The feature maps from a convolutional layer are fed through nonlinear activation functions. This makes it possible for the entire neural network to approximate almost any nonlinear function \cite{leshno1993multilayer, sonoda2017neural}\footnote{A neural network with only linear activations would only be able to perform linear approximation. Adding further layers wouldn't improve its expressiveness.} The activation functions are generally the very simple rectified linear units, or ReLUs, defined as  $\mbox{ReLU}(z) = \max(0, z)$, or variants like leaky ReLUs or parametric ReLUs.\footnote{Other options include exponential linear units (ELUs), and the now rarely used sigmoid or tanh activation functions.} See \cite{clevert2015fast, he2015delving} for more information about these and other activation functions. Feeding the feature maps through an activation function produces new tensors, typically also called \textit{feature maps}.
    \item \textbf{Pooling}: Each feature map produced by feeding the data through one or more convolutional layer is then typically pooled in a \textit{pooling layer}. Pooling operations take small grid regions as input and produce single numbers for each region. The number is usually computed by using the max function (\textit{max-pooling}) or the average function (\textit{average pooling}). Since a small shift of the input image results in small changes in the activation maps, the pooling layers gives the CNN some translational invariance. 
    
    A different way of getting the downsampling effect of pooling is to use convolutions with increased stride lengths. Removing the pooling layers simplifies the network architecture without necessarily sacrificing performance \cite{springenberg2014striving}.

\end{enumerate}

Other common elements in many modern CNNs include

\begin{enumerate}[iv)]
    \item \textbf{Dropout regularization}: A simple idea that gave a huge boost in the performance of CNNs. By averaging several models in an ensemble one tend to get better performance than when using single models. Dropout \cite{srivastava2014dropout} is an averaging technique based on stochastic sampling of neural networks.\footnote{The idea of dropout is also used for other machine learning models, as in the DART technique for regression trees \cite{rashmi2015dart}} By randomly removing neurons during training one ends up using slightly different networks for each batch of training data, and the weights of the trained network are tuned based on optimization of multiple variations of the network.\footnote{In addition to increased model performance, dropout can also be used to produce robust uncertainty measures in neural networks. By leaving dropout turned on also during inference one effectively performs \textit{variational inference} \cite{gal2016uncertainty, murray2018, wickstrom2018uncertainty}. This relates standard deep neural networks to Bayesian neural networks, synthesized in the field of Bayesian deep learning.}
    \item \textbf{Batch normalization}: These layers are typically placed after activation layers, producing normalized activation maps by subtracting the mean and dividing by the standard deviation for each training batch. Including batch normalization layers forces the network to periodically change its activations to zero mean and unit standard deviation as the training batch hits these layers, which works as a regularizer for the network, speeds up training, and makes it less dependent on careful parameter initialization \cite{ioffe2015batch}.
\end{enumerate}

In the design of new and improved CNN architectures, these components are combined in increasingly complicated and interconnected ways, or even replaced by other more convenient operations. When architecting a CNN for a particular task there are multiple factors to consider, including understanding the task to be solved and the requirements to be met, figuring out how to best feed the data to the network, and optimally utilizing one's budget for computation and memory consumption. In the early days of modern deep learning one tended to use very simple combinations of the building blocks, as in Lenet \cite{lecun1998gradient} and AlexNet \cite{NIPS2012_4824}. Later network architectures are much more complex, each generation building on ideas and insights from previous architectures, resulting in updates to the state-of-the-art. Table \ref{tab:CNNs} contains a short list of some famous CNN architectures, illustrating how the building blocks can be combined and how the field moves along.

\renewcommand{\arraystretch}{2.0}
{\footnotesize
\centering

\begin{longtable}{llp{11cm}} 
\caption{A far from exhaustive, non-chronological, list of CNN architectures and some high-level descriptions} \label{tab:CNNs} \\
AlexNet & \cite{NIPS2012_4824} & The network that launched the current deep learning boom by winning the 2012 ILSVRC competition by a huge margin. Notable features include the use of RELUs, dropout regularization, splitting the computations on multiple GPUs, and using data augmentation during training. ZFNet \cite{zeiler2014visualizing}, a relatively minor modification of AlexNet, won the 2013 ILSVRC competition. \\\hline
VGG & \cite{simonyan2014very} & Popularized the idea of using smaller filter kernels and therefore deeper networks (up to 19 layers for VGG19, compared to 7 for AlexNet and ZFNet), and training the deeper networks using pre-training on shallower versions. \\ \hline
GoogLeNet  & \cite{szegedy2015going} & Promoted the idea of stacking the layers in CNNs more creatively, as \textit{networks in networks}, building on the idea of \cite{lin2013network}. Inside a relatively standard architecture (called the \textit{stem}), GoogLeNet contains multiple \textit{inception modules}, in which multiple different filter sizes are applied to the input and their results concatenated. This multi-scale processing allows the module to extract features at different levels of detail simultaneously. GoogLeNet also popularized the idea of not using fully-connected layers at the end, but rather global average pooling, significantly reducing the number of model parameters. It won the 2014 ILSVRC competition.  \\ \hline
ResNet & \cite{he2016deep} & Introduced \textit{skip connections}, which makes it possible to train much deeper networks. A 152 layer deep ResNet won the 2015 ILSVRC competition, and the authors also successfully trained a version with \textit{1001} layers. Having skip connections in addition to the standard pathway gives the network the option to simply copy the activations from layer to layer (more precisely, from ResNet block to ResNet block), preserving information as data goes through the layers. Some features are best constructed in shallow networks, while others require more depth. The skip connections facilitate both at the same time, increasing the network's flexibility when fed input data. As the skip connections make the network learn residuals, ResNets perform a kind of boosting. \\ \hline
Highway nets &\cite{srivastava2015training} & Another way to increase depth  based on \textit{gating units}, an idea from Long Short Term Memory (LSTM) recurrent networks, enabling optimization of the skip connections in the network. The gates can be trained to find useful combinations of the identity function (as in ResNets) and the standard nonlinearity through which to feed its input.  \\\hline
DenseNet & \cite{huang2017densely} & Builds on the ideas of ResNet, but instead of adding the activations produced by one layer to later layers, they are simply concatenated together. The original inputs in addition to the activations from previous layers are therefore kept at each layer (again, more precisely, between blocks of layers), preserving some kind of global state. This encourages feature reuse and lowers the number of parameters for a given depth. DenseNets are therefore particularly well-suited for smaller data sets (outperforming others on e.g. Cifar-10 and Cifar-100). \\\hline
ResNext & \cite{xie2017aggregated} & Builds on ResNet and GoogLeNet by using inception modules between skip connections.\\ \hline
SENets & \cite{hu2017squeeze} & Squeeze-and-Excitation Networks, which won the ILSVRC 2017 competition, builds on ResNext but adds trainable parameters that the network can use to weigh each feature map, where earlier networks simply added them up. These SE-blocks allows the network to model the channel and spatial information separately, increasing the model capacity. SE-blocks can easily be added to any CNN model, with negligible increase in computational costs.  \\\hline
NASNet & \cite{zoph2017learning} & A CNN architecture designed by a neural network, beating all the previous human-designed networks at the ILSVRC competition. It was created using AutoML\footnote{\url{https://cloud.google.com/automl}}, Google Brain's reinforcement learning approach to architecture design \cite{pmlr-v70-bello17a}. A controller network (a recurrent neural network) proposes architectures aimed to perform at a specific level for a particular task, and by trial and error learns to propose better and better models. NASNet was based on Cifar-10, and has relatively modest computational demands, but still outperformed the previous state-of-the-art on ILSVRC data.  \\\hline
YOLO & \cite{redmon2016you} & Introduced a new, simplified way to do simultaneous object detection and classification in images. It uses a single CNN operating directly on the image and outputting bounding boxes and class probabilities. It incorporates several elements from the above networks, including inception modules and pretraining a smaller version of the network. It's fast enough to enable real-time processing\footnote{You can watch YOLO in action here \url{https://youtu.be/VOC3huqHrss}}. YOLO makes it easy to trade accuracy for speed by reducing the model size. YOLOv3-tiny was able to process images at over 200 frames per second on a standard benchmark data set, while still producing reasonable predictions.\\ \hline
GANs &\cite{NIPS2014_5423} & A generative adversarial network consists of two neural networks pitted against each other. The \textit{generative network} G is tasked with creating samples that the \textit{discriminative network D} is supposed to classify as coming from the generative network or the training data. The networks are trained simultaneously, where G aims to maximize the probability that D makes a mistake while D aims for high classification accuracy. \\\hline
Siamese nets & \cite{koch2015siamese}& An old idea (e.g. \cite{bromley1994signature}) that's recently been shown to enable \textit{one-shot learning}, i.e. learning from a single example. A siamese network consists of two identical neural networks, both the architecture and the weights, attached at the end. They are trained together to \textit{differentiate} pairs of inputs. Once trained, the features of the networks can be used to perform one-shot learning without retraining.   \\ \hline
U-net & \cite{ronneberger2015u} & A very popular and successful network for segmentation in 2D images. When fed an input image, it is first downsampled through a ``traditional" CNN, before being upsampled using transpose convolutions until it reaches its original size. In addition, based on the ideas of ResNet, there are skip connections that concatenates features from the downsampling to the upsampling paths. It is a fully-convolutional network, using the ideas first introduced in \cite{long2015fully}. \\\hline
V-net & \cite{milletari2016v} & A three-dimensional version of U-net with volumetric convolutions and skip-connections as in ResNet. \\
\end{longtable}

}

These neural networks are typically implemented in one or more of a small number of software frameworks that dominates machine learning research, all built on top of NVIDIA's CUDA platform and the cuDNN library. Today's deep learning methods are almost exclusively implemented in either TensorFlow, a framework originating from Google Research, Keras, a deep learning library originally built by Fran\c{c}ois Chollet and recently incorporated in TensorFlow, or Pytorch, a framework associated with Facebook Research. There are very few exceptions (YOLO built using the Darknet framework \cite{darknet13} is one of the rare ones). All the main frameworks are open source and under active development.

\section{Deep learning, medical imaging and MRI}
\label{S:3}

Deep learning methods are increasingly used to improve clinical practice, and the list of examples is long, growing daily. We will not attempt a comprehensive overview of deep learning in medical imaging, but merely sketch some of the landscape before going into a more systematic exposition of deep learning in MRI.

Convolutional neural networks can be used for efficiency improvement in radiology practices through protocol determination based on short-text classification \cite{Lee2018a}. They can also be used to reduce the gadolinium dose in contrast-enhanced brain MRI by an order of magnitude \cite{Gong2018} without significant reduction in image quality. Deep learning is applied in radiotherapy  \cite{Meyer2018},  in PET-MRI attenuation correction \cite{Liu2018,Mehranian2016},  in \textit{radiomics} \cite{Lao2017,Oakden-Rayner2017} (see \cite{Peeken2018} for a 
review of radiomics related to radiooncology and medical physics), and for {\it theranostics} in neurosurgical imaging, combining confocal laser endomicroscopy with deep learning models for automatic detection of intraoperative CLE images on-the-fly \cite{Izadyyazdanabadi2018}. 

Another important application area is advanced deformable image registration, enabling quantitative analysis across different physical imaging modalities and across time.\footnote{e.g. test-retest examinations, or motion correction in dynamic imaging}. For example elastic registration between 3D MRI and transrectal ultrasound for guiding targeted prostate biopsy 
\cite{Haskins2018}; deformable registration for brain MRI where a ``cue-aware deep regression network'' learns from a given set of training images the displacement vector associated with a pair of reference-subject patches \cite{Cao2018};
fast deformable image registration of brain MR image pairs by patch-wise prediction of the Large Deformation Diffeomorphic Metric Mapping model \cite{Yang2017}\footnote{available at \url{https://github.com/rkwitt/quicksilver}}; unsupervised convolutional neural network-based algorithm for deformable image registration of cone-beam CT to CT using a deep convolutional inverse graphics network \cite{Kearney2018}; deep learning-based 2D/3D registration framework for registration of preoperative 3D data and intraoperative 2D X-ray images in image-guided therapy \cite{Zheng2018};
real-time prostate segmentation during targeted prostate biopsy, utilizing temporal information in the series of ultrasound images \cite{Anas2018}.

This is just a tiny sliver of the many applications of deep learning to central problems in medical imaging. There are several thorough reviews and overviews of the field to consult for more information, across modalities and organs, and with different points of view and level of technical details. For example the comprehensive review \cite{Ching2018}\footnote{A continuous collaborative manuscript (\url{https://greenelab.github.io/deep-review}) with $>$500 references.}, covering both medicine and biology and spanning from imaging applications in healthcare to protein-protein interaction and uncertainty quantification;
key concepts of deep learning for clinical radiologists \cite{Lee2017,Rueckert2016,Chartrand2017,Erickson2017,Mazurowski2018,McBee2018,Savadjiev2018,Thrall2018,Yamashita2018,Yasaka2018}, including radiomics and imaging genomics (radiogenomics) \cite{Giger2018}, and 
toolkits and libraries for deep learning \cite{Erickson2017a};
deep learning in neuroimaging and neuroradiology \cite{Zaharchuk2018};
brain segmentation \cite{Akkus2017};
stroke imaging \cite{Lee2017b,Feng2018};
neuropsychiatric disorders \cite{Vieira2017};
breast cancer \cite{Burt2018,Samala2017}; 
chest imaging \cite{Ginneken2017};
imaging in oncology \cite{Morin2018,Parmar2018,Xue2017}; 
medical ultrasound \cite{Brattain2018,Huang2018};
and more technical surveys of deep learning in medical image analysis \cite{Litjens2017,Shen2017,Suzuki2017,Cao2018a}. Finally, for those who like to be hands-on, there are many instructive introductory deep learning tutorials available online. For example \cite{Lakhani2018}, with accompanying code available at \url{https://github.com/paras42/Hello_World_Deep_Learning}, where you'll be guided through the construction of a system that can differentiate a chest X-ray from an abdominal X-ray using the {\tt Keras}/{\tt TensorFlow} framework through a Jupyter Notebook. Other nice tutorials are \url{http://bit.ly/adltktutorial}, based on the Deep Learning Toolkit (DLTK) \cite{pawlowski2017state}, and \url{https://github.com/usuyama/pydata-medical-image}, based on the Microsoft Cognitive Toolkit (CNTK).\\

\begin{figure}[H]
\begin{tabular}{ll}
\hspace{-10mm} & \includegraphics[width=1\linewidth]{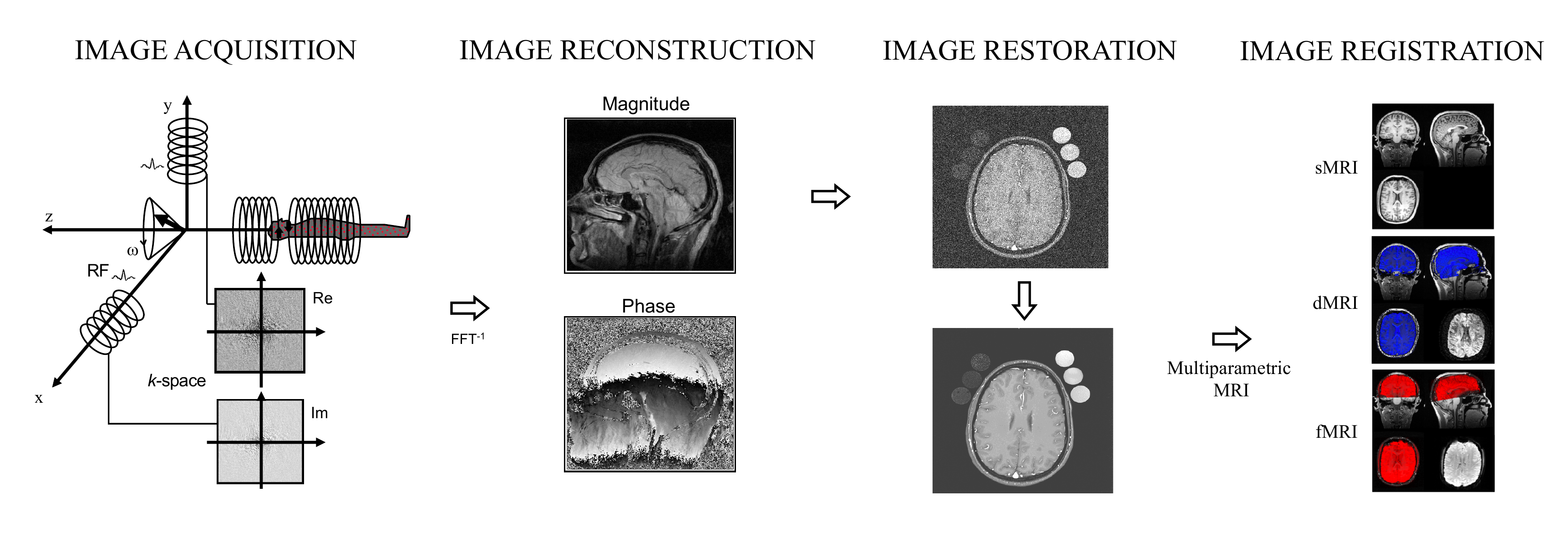}
\end{tabular}
\vspace{-6mm}
\caption{Deep learning in the MR signal processing chain, from image acquisition (in complex-valued $k$-space) and image reconstruction, to image restoration (e.g. denoising) and image registration. The rightmost column illustrates coregistration of multimodal brain MRI. sMRI = structural 3D T1-weighted MRI, dMRI = diffusion weighted MRI (stack of slices in blue superimposed on sMRI), fMRI = functional BOLD MRI (in red).}
\label{fig:acquisition}
\end{figure}

\begin{figure}[H]
\begin{tabular}{ll}
\hspace{-10mm} & \includegraphics[width=1\linewidth]{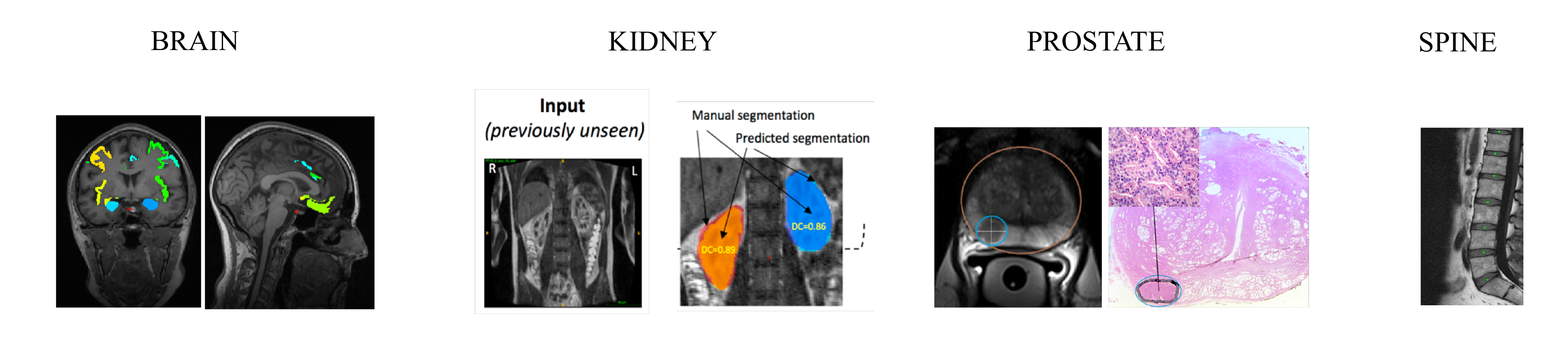}
\end{tabular}
\vspace{-8mm}
\caption{\it Deep learning for MR image analysis in selected organs, partly from ongoing work at MMIV.}
\label{fig:analysis}
\end{figure}

Let's now turn to the field of MRI, in which deep learning has seen applications at each step of entire workflows. From  acquisition  to  image retrieval, from segmentation to disease prediction. We divide this into two parts: ({\bf i}) the signal processing chain close to the physics of MRI, including image restoration and multimodal image registration (Fig.~\ref{fig:acquisition}), 
and ({\bf ii}) the use of deep learning in MR image segmentation, disease detection, disease prediction and systems based on images and text data (reports), addressing a few selected organs such as the brain, the kidney, the prostate and the spine (Fig.~\ref{fig:analysis}).\\

\vspace{-2mm}

\subsection{\bf From image acquisition to image registration}
\label{S:4.1}

Deep learning in MRI has typically been focused on segmentation and classification of reconstructed magnitude images. Its penetration into the lower levels of MRI measurement techniques is more recent, but already impressive. From MR image acquisition and signal processing in MR fingerprinting, to denoising and super-resolution, and into image synthesis.

\subsubsection{Data acquisition and image reconstruction}

Research on CNN and RNN-based image reconstruction methods is rapidly increasing, pioneered by Yang et al. \cite{Yang2016} at NIPS 2016 and Wang et al. \cite{wang2016accelerating} at ISBI 2016. Recent applications addresses e.g. convolutional recurrent neural networks for dynamic MR image reconstruction \cite{Qin2018}, reconstructing good quality cardiac MR images from highly undersampled complex-valued $k$-space data by learning spatio-temporal dependencies, outperforming 3D CNN approaches and compressed sensing-based dynamic MRI reconstruction algorithms in computational complexity, reconstruction accuracy and speed for different undersampling rates. Schlemper et.al. \cite{Schlemper2018} created a {\it deep cascade} of concatenated CNNs for dynamic MR image reconstruction, making use of data augmentation, both rigid and elastic deformations, to increase the variation of the examples seen by the network and reduce overfitting\footnote{Code available at \url{https://github.com/js3611/Deep-MRI-Reconstruction}}. Using \textit{variational networks} for single-shot fast spin-echo MRI with variable density sampling, Chen et.al. \cite{Chen2018} enabled real-time (200 ms per section) image reconstruction, outperforming conventional parallel imaging and compressed sensing reconstruction. In \cite{Knoll2018}, the authors explored the potential for {\it transfer learning} (pretrained models) and assessed the generalization of learned image reconstruction regarding image contrast, SNR, sampling pattern and image content, using a variational network and true measurement $k$-space data from patient knee MRI recordings and synthetic $k$-space data generated from images in the Berkeley Segmentation Data Set and Benchmarks. Employing least-squares generative adversarial networks (GANs) that learns texture details and suppresses high-frequency noise, \cite{Mardani2018} created a novel compressed sensing framework that can produce diagnostic quality reconstructions ``on the fly'' (30 ms)\footnote{In their GAN setting, a generator network is used to map undersampled data to a realistic-looking image with high measurement fidelity, while a discriminator network is trained jointly to score the quality of the reconstructed image.}. A unified framework for image reconstruction \cite{Zhu2018}, called \textit{automated transform by manifold approximation} (AUTOMAP) consisting of a feedforward deep neural network with fully connected layers followed by a sparse convolutional autoencoder, formulate image reconstruction generically as a data-driven supervised learning task that generates a mapping between the sensor and the image domain based on an appropriate collection of training data (e.g. MRI examinations collected from the Human Connectome Project, transformed to the $k$-space sensor domain).\\

There are also other approaches and reports on deep learning in MR image reconstruction, e.g. \cite{Eo2018,Han2018,Shi2018,Yang2018}, a fundamental field rapidly progressing.

\subsubsection{Quantitative parameters - QSM and MR fingerprinting}

Another area that is developing within deep learning for MRI is the estimation of quantitative tissue parameters from recorded complex-valued data. For example within \textit{quantitative susceptibility mapping}, and in the exciting field of \textit{magnetic resonance fingerprinting}.

Quantitative susceptibility mapping (QSM) is a growing field of research in MRI, aiming to noninvasively estimate the magnetic susceptibility of biological tissue \cite{deistung2013toward, deistung2017overview}. The technique is based on solving the difficult, ill-posed inverse problem of determining the magnetic susceptibility from local magnetic fields. Recently Yoon et al. \cite{Yoon2018} constructed a three-dimensional CNN, named QSMnet and based on the U-Net architecture, able to generate high quality susceptibility source maps from single orientation data. The authors generated training data by using the gold-standard for QSM: the so-called COSMOS method \cite{liu2009calculation}. The data was based on 60 scans from 12 healthy volunteers. The resulting model both simplified and improved the state-of-the-art for QSM. Rasmussen and coworkers \cite{rasmussen2018deepqsm} took a different approach. They also used a U-Net-based convolutional neural network to perform field-to-source inversion, called \textit{DeepQSM}, but it was trained on synthetically generated data containing simple geometric shapes such as cubes, rectangles and spheres. After training their model on synthetic data it was able to generalize to real-world clinical brain MRI data, computing susceptibility maps within seconds end-to-end. The authors conclude that their method, combined with fast imaging sequences, could make QSM feasible in standard clinical practice.

Magnetic resonance fingerprinting (MRF) was introduced a little more than five years ago \cite{Ma2013}, and has been called  ``a promising new approach to obtain standardized imaging biomarkers from MRI" by the European Society of Radiology \cite{RadiologyESR2015}. It uses a pseudo-randomized acquisition that causes the signals from different tissues to have a unique signal evolution (``fingerprint'') that is a function of the multiple material properties being investigated. Mapping the signals back to known tissue parameters (T1, T2 and proton density) is then a rather difficult inverse problem. MRF is closely related to the idea of compressed sensing \cite{Donoho2006} in MRI \cite{Lustig2007} in that MRF undersamples data in $k$-space producing aliasing artifacts in the reconstructed images that can be suppressed by compressed sensing.\footnote{See \cite{mccann2017convolutional, shah2018solving, lucas2018using, aggarwal2018modl, li2018nett} for recent perspectives and developments connecting deep learning-based reconstruction methods to the more general research field of inverse problems.} It can be regarded as a quantitative multiparametric MRI analysis, and with recent acquisition schemes using a single-shot spiral trajectory with undersampling, whole-brain coverage of T$_1$, T$_2$ and proton density maps can be acquired at $1.2 \times 1.2 \times 3$ mm$^3$ voxel resolution in less than 5 min \cite{Ma2018}.

The processing of MRF after acquisition usually involves using various pattern recognition algorithms that try to match the fingerprints to a predefined dictionary of predicted signal evolutions\footnote{A dictionary of time series for every possible combination of {\it parameters} like (discretized) T$_1$ and T$_2$ relaxation times, spin-density (M$_0$), B$_0$, off-resonance ($\Delta f$), and also voxel-wise cerebral blood volume (CBV), mean vessel radius (R), blood oxygen saturation (SO$_2$) and T$_2^*$ \cite{Christen2014,Lemasson2016,Rieger2018}, and more, e.g. MFR-ASL \cite{Wright2018}.}, created using the Bloch equations \cite{Ma2013,Panda2017}. 

Recently, deep learning methodology has been applied to MR fingerprinting. Cohen et al. \cite{Cohen2018} reformulated the MRF reconstruction problem as learning an optimal function that maps the recorded signal magnitudes to the corresponding tissue parameter values, trained on a sparse set of dictionary entries. To achieve this they fed voxel-wise MRI data acquired with an MRF sequence (MRF-EPI, 25 frames in $\sim$3 s; or MRF-FISP, 600 frames in $\sim$7.5 s) to a four-layer neural network consisting of two hidden layers with $300\times 300$ fully connected nodes and two nodes in the output layer, considering only T$_1$ and T$_2$ parametric maps. The network, called MRF Deep RecOnstruction NEtwork (DRONE), was trained by an adaptive moment estimation stochastic gradient descent algorithm with a mean squared error loss function. Their dictionary consisted of $\sim$70000 entries (product of discretized T$_1$ and T$_2$ values) and training the network to convergence with this dictionary ($\sim$10~MB for MRF-EPI  and $\sim$300~MB for MRF-FISP) required 10 to 70 min using an NVIDIA K80 GPU with 2 GB memory.  They found their reconstruction time (10 to 70 ms per slice) to be 300 to 5000 times faster than conventional dictionary-matching techniques, using both well-characterized calibrated ISMRM/NIST phantoms and in vivo human brains.

A similar deep learning approach to predict quantitative parameter values (T$_1$ and T$_2$) from MRF time series was taken by Hoppe et al. \cite{Hoppe2017}. In their experiments they used 2D MRF-FISP data with variable TR (12-15 ms), flip angles (5$^{\circ}$-74$^{\circ}$) and 3000 repetitions, recorded on a MAGNETOM 3T Skyra. A high resolution dictionary was simulated to generate a large collection of training and testing data, using tissues T$_1$ and T$_2$ relaxation time ranges as present in normal brain at 3T (e.g. \cite{Bojorquez2017}) resulting in $\sim1.2\times 10^5$ time series. In contrast to \cite{Cohen2018}, their deep neural network architecture was inspired from the domain of speech recognition due to the similarity of the two tasks. The architecture with the smallest average error for validation data was a standard convolutional neural network consisting of an input layer of 3000 nodes (number of samples in the recorded time series), four hidden layers, and an output layers with two nodes (T$_1$ and T$_2$). Matching one time series was about 100 times faster than the conventional \cite{Ma2013} matching method and with very small mean absolute deviations from ground truth values. 

In the same context, Fang et al. \cite{Fang2017} used a deep learning method to extract tissue properties from highly undersampled 2D MRF-FISP data in brain imaging, where 2300 time points were acquired from each measurement and each time point consisted of data from one spiral readout only. The real and imaginary parts of the complex signal were separated into two channels. They used MRF signal from a patch of $32\times32$ pixels to incorporate correlated information between neighboring pixels. In their work they designed a standard three-layer CNN with T$_1$ and T$_2$ as output. 

Virtue et.al. \cite{Virtue2017} investigated a different approach to MRF. By generating 100.000 synthetic MRI signals using a Bloch equation simulator they were able to train feedforward deep neural networks to map new MRI signals to the tissue parameters directly, producing approximate solutions to the inverse mapping problem of MRF. In their work they designed a new complex activation function, the complex cardioid, that was used to construct a complex-valued feedforward neural network. This three-layer network outperformed both the standard MRF techniques based on dictionary matching, and also the analogous real neural network operating on the real and imaginary components separately. This suggested that complex-valued networks are better suited at uncovering information in complex data.\footnote{Complex-valued deep learning is also getting some attention in a broader community of researchers, and has been shown to lead to improved models. See e.g. \cite{tygert2016mathematical, trabelsi2017deep} and the references therein.}

\subsubsection{Image restoration (denoising, artifact detection)}

\noindent Estimation of noise and image denoising in MRI has been an important field of research for many years \cite{Sijbers1998,McVeigh1985}, employing a plethora of methods. For example Bayesian Markov random field models \cite{Baselice2017}, rough set theory \cite{Phophalia2017}, higher-order singular value decomposition \cite{Zhang2015}, wavelets \cite{VanDeVille2007}, independent component analysis \cite{Salimi-Khorshidi2014}, or higher order PDEs \cite{Lysaker2003}. 

Recently, deep learning approaches have been introduced to denoising. In their work on learning implicit brain MRI manifolds using deep neural networks, Bermudez et al. \cite{Bermudez2018} implemented an autoencoder with skip connections for image denoising, testing their approach with adding various levels of Gaussian noise to more than 500 T1-weighted brain MR images from healthy controls in the Baltimore Longitudinal Study of Aging. Their autoencoder network outperformed the current FSL SUSAN denoising software according to peak signal-to-noise ratios. Benou et al. \cite{Benou2017} addressed spatio-temporal denoising of dynamic contrast-enhanced MRI of the brain with bolus injection of contrast agent (CA), proposing a novel approach using ensembles of deep neural networks for noise reduction. Each DNN was trained on a different range of SNRs and types of CA concentration time curves (denoted ``pathology experts", ``healthy experts", ``vessel experts") to generate a reconstruction hypothesis from noisy input by using a classification DNN to select the most likely hypothesis and provide a ``clean output" curve. Training data was generated synthetically using a three-parameter Tofts pharmacokinetic (PK) model and noise realizations. To improve this model, accounting for spatial dependencies of PK pharmacokinetics, they used concatenated noisy time curves from first-order neighbourhood pixels in their expert DNNs and ensemble hypothesis DNN, collecting neighboring reconstructions before a boosting procedure produced the final clean output for the pixel of interest. They tested their trained ensemble model on 33 patients from two different DCE-MRI databases with either stroke or recurrent glioblastoma (RIDER NEURO\footnote{\url{https://wiki.cancerimagingarchive.net/display/Public/RIDER+NEURO+MRI}}), acquired at different sites, with different imaging protocols, and with different scanner vendors and field strengths. The qualitative and quantitative (MSE) denoising results were better than spatio-temporal Beltrami, moving average, the dynamic Non Local Means method \cite{Gal2010}, and stacked denoising autoencoders \cite{Vincent2010}. The run-time comparisons were also in favor of the proposed sDNN. 

In this context of DCE-MRI, it's tempting to speculate whether deep neural network approaches could be used for {\it direct estimation} of tracer-kinetic parameter maps from highly undersampled $({\bf k},t)$-space data in dynamic recordings \cite{Dikaios2014,Guo2017}, a powerful way to by-pass 4D DCE-MRI reconstruction altogether and map sensor data directly to spatially resolved pharmacokinetic parameters, e.g. K$^{trans}$, $v_p$, $v_e$ in the extended Tofts model or parameters in other classic models \cite{Sourbron2013}. A related approach in the domain of diffusion MRI, by-passing the model-fitting steps and computing voxel-wise scalar tissue properties (e.g. radial kurtosis, fiber orientation dispersion index) directly from the subsampled DWIs was  taken by Golkov et al. \cite{Golkov2016} in their proposed ``$q$-space deep learning" family of methods.\\

\vspace{-2mm}

Deep learning methods has also been applied to MR {\it artifact detection}, e.g. poor quality spectra in MRSI \cite{Gurbani2018}; detection and removal of ghosting artifacts in MR spectroscopy \cite{Kyathanahally2018}; and automated reference-free detection of patient motion artifacts in MRI \cite{Kuestner2018}.

\subsubsection{Image super-resolution}
\noindent Image super-resolution, reconstructing a higher-resolution image or image sequence from the observed low-resolution image \cite{Yue2016}, is an exciting application of deep learning methods\footnote{See \url{http://course.fast.ai/lessons/lesson14.html} for an instructive introduction to super-resolution}.

Super-resolution for MRI have been around for almost 10 years \cite{Shilling2009,Ropele2010} and can be used to improve the trade-off between resolution, SNR, and acquisition time \cite{Plenge2012}, generate 7T-like MR images on 3T MRI scanners \cite{Bahrami2017}, or obtain super-resolution T$_1$ maps from a set of low resolution T$_1$ weighted images \cite{VanSteenkiste2017}. 
Recently deep learning approaches has been introduced, e.g.
generating super-resolution single (no reference information) and multi-contrast (applying a high-resolution image of another modality as reference) brain MR images using CNNs \cite{Zeng2018a};
constructing super-resolution brain MRI by a CNN stacked by multi-scale fusion units \cite{Liu2018d}; and
super-resolution musculoskeletal MRI (``DeepResolve") \cite{Chaudhari2018}. In DeepResolve thin (0.7 mm) slices in knee images (DESS) from 124 patients included in the Osteoarthritis Initiative were used for training and 17 patients for testing, with a 10s inference time per 3D ($344 \times 344 \times 160$) volume. The resulting images were evaluated both quantitatively (MSE, PSNR, and the perceptual window-based structural similarity SSIM\footnote{\url{http://www.cns.nyu.edu/~lcv/ssim}} index) and qualitatively by expert radiologists.

\subsubsection{Image synthesis}

\noindent Image synthesis in MRI have traditionally been seen as a method to derive new parametric images or new tissue contrast from a collection of MR acquisition performed at the same imaging session, i.e. ``an intensity transformation applied to a given set of input images to generate a new image with a specific tissue contrast" \cite{Jog2017}. Another avenue of MRI synthesis is related to
quantitative imaging and the development and use of physical phantoms, imaging calibration/standard test objects with specific material properties. This is done in order to assess the performance of an MRI scanner or to assess imaging biomarkers reliably with application-specific phantoms such as a structural brain imaging phantom, DCE-MRI perfusion phantom, diffusion phantom, flow phantom, breast phantom or a proton-density fat fraction phantom \cite{Keenan2018}. The {\it in silico} modeling of MR images with certain underlying properties, e.g. \cite{Jurczuk2014,Zhou2018}, or model-based generation of large databases of (cardiac) images from real healthy cases \cite{Duchateau2018} is also part of this endeavour. In this context, deep learning approaches have accelerated research and the amount of costly training data. \\

\vspace{-2mm}

The last couple of years have seen impressive results for photo-realistic image synthesis using deep learning techniques, especially generative adversarial networks (GANs, introduced by Goodfellow et al. in 2014 \cite{NIPS2014_5423}), e.g.  \cite{Creswell2018,Hong2017,Huang2018a}. These can also be used for biological image synthesis \cite{Osokin2017,Antipov2017} and text-to-image synthesis \cite{Bodnar2018,Dong2017a,Reed2016}.\footnote{See here \url{https://github.com/xinario/awesome-gan-for-medical-imaging} for a list of interesting applications of GAN in medical imaging}
Recently, a group of researchers from NVIDIA, MGH \& BWH Center for Clinical Data Science in Boston, and the Mayo Clinic in Rochester \cite{shin2018medical} designed a clever approach to generate synthetic abnormal MRI images with brain tumors by training a GAN based on {\tt pix2pix}\footnote{\url{https://phillipi.github.io/pix2pix}} using two publicly available data sets of brain MRI (ADNI and the BRATS'15 Challenge, and later also the Ischemic Stroke Lesion Segmentation ISLES'2018 Challenge). This approach is highly interesting as medical imaging datasets are often imbalanced, with few pathological findings, limiting the training of deep learning models. Such generative models for image synthesis serve as a form of data augmentation, and also as an anonymization tool. The authors achieved comparable tumor segmentation results when trained on the synthetic data rather than on real patient data. A related approach to brain tumor segmentation using coarse-to-fine GANs was taken by Mok \& Chung \cite{Mok2018}.
Guibas et al. \cite{Guibas2017} used a two-stage pipeline for generating synthetic medical images from a pair of GANs, addressing retinal fundus images, and provided an 
online repository (SynthMed) for synthetic medical images. Kitchen \& Seah \cite{Kitchen2017} used GANs to synthetize realistic prostate lesions in T$_2$, ADC, K$^{trans}$ resembling the SPIE-AAPM-NCI ProstateX Challenge 2016\footnote{\url{https://www.aapm.org/GrandChallenge/PROSTATEx-2}} training data.

Other applications are unsupervised synthesis of T1-weighted brain MRI using a GAN  \cite{Bermudez2018};
image synthesis with context-aware GANs \cite{Nie2017};
synthesis of patient-specific transmission image for PET attenuation correction in PET/MR imaging of the brain using a CNN \cite{Spuhler2018};
pseudo-CT synthesis for pelvis PET/MR attenuation correction using a Dixon-VIBE Deep Learning (DIVIDE) network \cite{Torrado-Carvajal2018};
image synthesis with GANs for tissue recognition \cite{Zhang2018};
synthetic data augmentation using a GAN for improved liver lesion classification \cite{Frid-Adar2018}; and
deep MR to CT synthesis using unpaired data \cite{Wolterink2017}.

\subsubsection{Image registration}

\noindent Image registration\footnote{Image registration can be defined as ``the determination of a one-to-one mapping between the coordinates in one space and those in another, such that points in the two spaces that correspond to the same anatomical point are mapped to each other" (C.R Maurer \cite{Maurer1993}, 1993).} is an increasingly important field within MR image processing and analysis as more and more complementary and multiparametric tissue information are collected in space and time within shorter acquisition times, at higher spatial (and temporal) resolutions, often longitudinally, and across patient groups, larger cohorts, or atlases. Traditionally one has divided the tasks of image registration into dichotomies: intra vs. inter-modality, intra vs. inter-subject, rigid vs. deformable, geometry-based vs. intensity-based, and prospective vs. retrospective image registration. Mathematically, registration is a challenging mix of geometry (spatial transformations), analysis (similarity measures), optimization strategies, and numerical schemes. In prospective motion correction, real-time MR physics is also an important part of the picture \cite{Maclaren2013,Zaitsev2017}.  A wide range of methodological approaches have been developed and tested for various organs and applications\footnote{and different hardware e.g. GPUs \cite{Fluck2011,Shi2012,Eklund2013} as image registration is often computationally time consuming.} \cite{Maintz1998,Glocker2011,Sotiras2013,Oliveira2014,Saha2015,Viergever2016,Song2017,Ferrante2017,Keszei2017,Nag2017}, including ``previous generation" artificial neural networks \cite{Jiang2010}.

Recently, deep learning methods have been applied to image registration in order to improve accuracy and speed (e.g. Section 3.4 in \cite{Litjens2017}). For example: 
deformable image registration \cite{Wu2016,Yang2017};
model-to-image registration \cite{Salehi2018,Toth2018};
MRI-based attenuation correction for PET \cite{Han2017,Liu2018c};
PET/MRI dose calculation \cite{Xiang2017};
unsupervised end-to-end learning for deformable registration of 2D CT/MR images \cite{Shan2017};
an unsupervised learning model for deformable, pairwise 3D medical image registration by  Balakrishnan et al. \cite{Balakrishnan2018}\footnote{with code available at \url{https://github.com/voxelmorph/voxelmorph}}; and
a deep learning framework for unsupervised affine and deformable image registration \cite{Vos2018}.

\vspace{5mm}

\subsection{\bf From image segmentation to diagnosis and prediction}
\label{S:4.2}

We leave the lower-level applications of deep learning in MRI to consider higher-level (downstream) applications such as fast and accurate image segmentation, disease prediction in selected organs (brain, kidney, prostate, and spine) and content-based image retrieval, typically applied to reconstructed magnitude images. We have chosen to focus our overview on deep learning applications close to the MR physics and will be brief in the present section, even if the following applications are very interesting and clinically important.

\subsubsection{Image segmentation}

\noindent Image segmentation, the holy grail of quantitative image analysis, is the process of partitioning an image into multiple regions that share similar attributes, enabling localization and quantification.\footnote{Segmentation is also crucial for functional imaging, enabling tissue physiology quantification with preservation of anatomical specificity.
}
It has an almost 50 years long history, and has become the biggest target for deep learning approaches in medical imaging. The multispectral tissue classification report by Vannier et al. in 1985 \cite{Vannier1985}, using statistical pattern recognition techniques (and satellite image processing software from NASA), represented one of the most seminal works leading up to today's machine learning in medical imaging segmentation. In this early era, we also had the opportunity to contribute with supervised and unsupervised machine learning approaches for MR image segmentation and tissue classification \cite{Lundervold1988,Taxt1992,Taxt1994,Lundervold1995}. 
An impressive range of segmentation methods and approaches have been reported (especially for brain segmentation) and reviewed, e.g. \cite{Cabezas2011,Garcia-Lorenzo2013,Smistad2015,Bernal2017,Dora2017,Torres2018,Bernal2018,Moccia2018,Makropoulos2018}.
MR image segmentation using deep learning approaches, typically CNNs, are now penetrating the whole field of applications. For example acute ischemic lesion segmentation in DWI \cite{Chen2017}; brain tumor segmentation \cite{Havaei2017}; segmentation of the striatum \cite{Choi2016}; 
segmentation of organs-at-risks in head and neck CT images \cite{Ibragimov2017}; and fully automated segmentation of polycystic kidneys \cite{Kline2017}; deformable segmentation of the prostate \cite{Guo2016}; and spine segmentation with 3D multiscale CNNs \cite{Li2018c}.

See \cite{Litjens2017} and \cite{Ching2018} for more comprehensive lists. \\

\subsubsection{Diagnosis and prediction}

\noindent A presumably complete list of papers up to 2017 using deep learning techniques for brain image analysis is provided as Table 1 in Litjens at al. \cite{Litjens2017}. In the following we add some more recent work on organ-specific deep learning using MRI, restricting ourselves to brain, kidney, prostate and spine.

\renewcommand{\arraystretch}{2.0}
{\footnotesize
\centering

\begin{longtable}{p{3.5cm}p{0.5cm}p{10.0cm}} 

\caption{\small A short list of deep learning applications per {\it organ}, {\it task}, {\it reference} and {\it description}.} 
\label{tab:diagnosis} \\
\hline\\
\textbf{BRAIN} &   &   \\ \hline \hline
Brain extraction  &  \cite{Kleesiek2016} & A 3D CNN for skull stripping \\ 
Functional connectomes & \cite{Li2018a} & Transfer learning approach to enhance deep neural network classification of brain functional connectomes  \\ 
                          & \cite{Zeng2018} & Multisite diagnostic classification of schizophrenia using discriminant deep learning with functional connectivity MRI \\ 
Structural connectomes & \cite{Wasserthal2018} & A convolutional neural network-based approach (\url{https://github.com/MIC-DKFZ/TractSeg}) that directly segments tracts in the field of fiber orientation distribution function (fODF) peaks without using tractography, image registration or parcellation. Tested on 105 subjects from the Human Connectome Project \\ 
Brain age &  \cite{Cole2017} & Chronological age prediction from raw brain T1-MRI data,  also testing the heritability of brain-predicted age using a sample of 62 monozygotic and dizygotic twins\\
Alzheimer's disease &  \cite{Liu2018f} & Landmark-based deep multi-instance learning evaluated on 1526 subjects from three public datasets (ADNI-1, ADNI-2, MIRIAD) \\ 
                    &  \cite{Islam2018} &  Identify different stages of AD  \\ 
                    &  \cite{Lu2018} & Multimodal and multiscale deep neural networks for the early diagnosis of AD using structural MR and FDG-PET images \\ 
Vascular lesions &  \cite{Moeskops2018} & Evaluation of a deep learning approach for the segmentation of brain tissues and white matter hyperintensities of presumed vascular origin in MRI \\ 
Identification of MRI contrast &  \cite{Pizarro2018} & Using deep learning algorithms to automatically identify the brain MRI contrast, with implications for managing large databases \\ 
Meningioma & \cite{Laukamp2018} & Fully automated detection and segmentation of meningiomas using deep learning on routine multiparametric MRI \\ 
Glioma  & \cite{Perkuhn2018} & Glioblastoma segmentation using heterogeneous MRI data from clinical routine \\ 
    & \cite{AlBadawy2018} & Deep learning for segmentation of brain tumors and impact of cross-institutional training and testing\\ 
    & \cite{Cui2018} & Automatic semantic segmentation of brain gliomas from MRI using a deep cascaded neural network \\ 
    & \cite{Hoseini2018} & AdaptAhead optimization algorithm for learning deep CNN applied to MRI segmentation of glioblastomas (BRATS) \\ 
Multiple sclerosis & \cite{Yoo2018} &  Deep learning of joint myelin and T1w MRI features in normal-appearing brain tissue to distinguish between multiple sclerosis patients and healthy controls \\ 
%          &   & \\
{\bf KIDNEY} &   &   \\ \hline \hline
Abdominal organs & \cite{Bobo2018} & CNNs to improve abdominal organ segmentation, including  left kidney, right kidney, liver, spleen,  and stomach in T$_2$-weighted MR images \\
Cyst segmentation & \cite{Kline2017} & An artificial multi-observer deep neural network for fully automated segmentation of polycystic kidneys \\ 
Renal transplant & \cite{Shehata2018} & A deep-learning-based classifier with stacked non-negative constrained autoencoders to distinguish between rejected and non-rejected renal transplants in DWI recordings\\ 
%          &   & \\
{\bf PROSTATE}    &  &   \\ \hline \hline
Cancer (PCa) & \cite{Cheng2017}   & Proposed a method for end-to-end prostate segmentation by integrating holistically (image-to-image) nested edge detection with fully convolutional networks. their nested networks automatically learn a hierarchical representation that can improve prostate boundary detection. Obtained very good results (Dice coefficient, 5-fold cross validation) on MRI scans from 250 patients \\ 
       & \cite{Ishioka2018} &  Computer-aided diagnosis with a CNN, deciding `cancer' `no cancer' trained on data from 301 patients with a prostate-specific antigen level of $<20$ ng/mL who underwent MRI and extended systematic prostate biopsy with or without MRI-targeted biopsy \\ 
       & \cite{Song2018}    & Automatic approach based on deep CNN, inspired from VGG, to classify PCa and noncancerous tissues with multiparametric MRI using data from the PROSTATEx database  \\
       & \cite{Wang2017}    & Deep CNN and a non-deep learning using feature detection (the scale-invariant feature transform and the bag-of-words model, a representative method for image recognition and analysis) were used to distinguish pathologically confirmed PCa patients from prostate benign conditions patients with prostatitis or prostate benign hyperplasia in a collection of 172 patients with more than 2500 morphologic 2D T$_2$-w MR images \\ 
       & \cite{Yang2017a} & Designed a system which can concurrently identify the presence of PCa in an image and localize lesions based on deep CNN features (co-trained CNNs consisting of two parallel convolutional networks for ADC and T$_2$-w images respectively) and a single-stage SVM classifier for automated detection of PCa in multiparametric MRI. Evaluated on a dataset of 160 patients \\ 
       & \cite{Le2017} & Designed and tested multimodel CNNs, using clinical data from 364 patients with a total of 463 PCa lesions and 450 identified noncancerous image patches. Carefully investigated three critical factors which could greatly affect the performance of their multimodal CNNs but had not been carefully studied previously: (1) Given limited training data, how can these be augmented in sufficient numbers and variety for fine-tuning deep CNN networks for PCa diagnosis? (2) How can multimodal mp-MRI information be effectively combined in CNNs? (3) What is the impact of different CNN architectures on the accuracy of PCa diagnosis?  \\ 
%          &   & \\
 {\bf SPINE} &    &   \\ \hline  \hline    
 Vertebrae labeling &  \cite{Forsberg2017} & Designed a CNN for detection and labeling of vertebrae in MR images with clinical annotations as training data  \\
 Intervertebral disc localization & \cite{Li2018c} & 3D multi-scale fully connected CNNs with random modality voxel dropout learning for intervertebral disc localization and segmentation from multi-modality MR images \\ 
 Disc-level labeling, spinal stenosis grading &  \cite{Lu2018a} & CNN model denoted DeepSPINE, having a U-Net architecture combined with a spine-curve fitting method for automated lumbar vertebral segmentation, disc-level designation, and spinal stenosis grading with a natural language processing scheme \\ 
 Lumbal neural forminal stenosis (LNFS) & \cite{Han2018d} & Addressed the challenge of automated pathogenesis-based diagnosis, simultaneously localizing and grading multiple spinal structures (neural foramina, vertebrae, intervertebral discs) for diagnosing LNFS and discover pathogenic factors. Proposed a deep multiscale multitask learning network integrating a multiscale multi-output learning and a multitask regression learning into a fully convolutional network where (i) a DMML-Net merges semantic representations to reinforce the salience of numerous target organs (ii) a DMML-Net extends multiscale convolutional layers as multiple output layers to boost the scale-invariance for various organs, and (iii) a DMML-Net joins the multitask regression module and the multitask loss module to combine the mutual benefit between tasks \\
Spondylitis vs tuberculosis & \cite{Kim2018} & CNN model for differentiating between tuberculous and pyogenic spondylitis in MR images. Compared their CNN performance with that of three skilled radiologists using spine MRIs from 80 patients \\ 
Metastasis & \cite{Wang2017} & A multi-resolution approach for spinal metastasis detection using deep Siamese neural networks comprising three identical subnetworks for multi-resolution analysis and detection. Detection performance was evaluated on a set of 26 cases using a free-response receiver operating characteristic analysis (observer is free to mark and rate as many suspicious regions as are considered clinically reportable)  \\ \hline
\end{longtable}

} % end footnotesize

\vspace{5mm}

\subsection{\bf Content-based image retrieval}

The objective of content-based image retrieval (CBIR) in radiology is to provide medical cases similar to a given image in order to assist radiologists in the decision-making process. It typically involves large case databases, clever image representations and lesion annotations, and algorithms that are able to quickly and reliably match and retrieve the most similar images and their annotations in the case database. CBIR has been an active area of research in medical imaging for many years, addressing a wide range of applications, imaging modalities, organs, and methodological approaches, e.g. \cite{Pilevar2011,Kumar2013,Faria2015,Kumar2015,Bedo2016,Muramatsu2018,Spanier2018}, and at a larger scale outside the medical field using deep learning techniques, e.g. at Microsoft,
Apple, Facebook, and Google (reverse image search\footnote{See ``search by image" \url{https://images.google.com}, \url{https://developers.google.com/custom-search}, and also \url{https://tineye.com}, indexing more than 30 billion images}), and others. See e.g. \cite{Gordo2016,Liu2017,Han2018c,Piplani2018,Yang2018a} and the code repositories {\small \url{https://github.com/topics/image-retrieval}}.
One of the first application of deep learning for CBIR in the medical domain came in 2015 when Sklan et al.  \cite{Sklan2015} trained a CNN to perform CBIR with more than one million random MR and CT images, with disappointing results (true positive rate of 20\%) on their independent test set of 2100 labeled images. Medical CBIR is now, however, dominated by deep learning algorithms \cite{Bressan2018,Qayyum2017,Chung2017}.
As an example, by retrieving medical cases similar to a given image, Pizarro et al. \cite{Pizarro2018} developed a CNN for automatically inferring the contrast of MRI scans based on the image intensity of multiple slices.\\

\vspace{-2mm}

\noindent Recently, deep learning methods have also been used for {\it automated generation of radiology reports}, typically incorporating long-short-term-memory (LSTM) network models to generate the textual paragraphs \cite{Jing2017,Li2018b,Moradi2018,Zhang2018a}, and also to identify findings in radiology reports \cite{Pons2016,Zech2018,Goff2018}.

\section{Open science and reproducible research in machine learning for medical imaging}
Machine learning is moving at a breakneck speed, too fast for the standard peer-review process to keep up. Many of the most celebrated and impactful papers in machine learning over the past few years are only available as preprints, or published in conference proceedings long after their results are well-known and incorporated in the research of others. Bypassing peer-review has some downsides, of course, but these are somewhat mitigated by researchers' willingness to share code and data.\footnote{In the spirit of sharing and open science, we've created a GitHub repository to accompany our article, available at \url{https://github.com/MMIV-ML/DLMI2018}.} 

Most of the main new ideas and methods are posted to the arXiv preprint server\footnote{\url{http://arxiv.org}}, and the accompanying code shared on the GitHub platform\footnote{\url{https://github.com}}. The data sets used are often openly available through various repositories. This, in addition to the many excellent online educational resources\footnote{For example  \url{http://www.fast.ai}, \url{https://www.deeplearning.ai}, \url{http://cs231n.stanford.edu}, \url{https://developers.google.com/machine-learning/crash-course}}, makes it easy to get started in the field. Select a problem you find interesting based on openly available data, a method described in a preprint, and an implementation uploaded to GitHub. This forms a good starting point for an interesting machine learning project.

Another interesting aspect about modern machine learning and data science is the prevalence of \textit{competitions}, with the now annual ImageNet Large Scale Visual Recognition Challenge (ILSVRC) competition as the main driver of progress in deep learning for computer vision since 2012. Each competition typically draws large number of participants, and the top results often push the state-of-the art to a new level. In addition to inspiring new ideas, competitions also provide natural entry points to modern machine learning. It is interesting to note how deep learning-based models are completely dominating the leaderboards of essentially all image-based competitions. Other machine learning models, or non-machine learning-based techniques, have largely been outclassed.

What's true about the openness of machine learning in general is increasingly true also for the sub-field of machine learning for medical image analysis. We've listed a few examples of openly available implementations, data sets and challenges in tables \ref{tab:code}, \ref{tab:data} and \ref{tab:competitions} below.

\renewcommand{\arraystretch}{2.0}
{\footnotesize
\centering

\begin{longtable}{|p{6cm}|p{2cm}|p{6cm}|} 
\caption{A short list of openly available code for ML in medical imaging} \label{tab:code} \\
%\hline\\
\textbf{Summary} & \textbf{Reference} & \textbf{Implementation}\\\hline
NiftyNet. An open source convolutional neural networks platform for medical image analysis and image-guided therapy & \cite{Gibson2018, Li17} & \url{http://niftynet.io}\\\hline
DLTK. State of the art reference implementations for deep learning on medical images & \cite{pawlowski2017state} & \url{https://github.com/DLTK/DLTK}\\\hline
DeepMedic & \cite{kamnitsas2017efficient} & \url{https://github.com/Kamnitsask/deepmedic}\\\hline
U-Net: Convolutional Networks for Biomedical Image Segmentation & \cite{RFB15a} & \url{https://lmb.informatik.uni-freiburg.de/people/ronneber/u-net}\\ \hline
V-net & \cite{milletari2016v} & \url{https://github.com/faustomilletari/VNet}\\\hline
SegNet: A Deep Convolutional Encoder-Decoder Architecture for Robust Semantic Pixel-Wise Labelling & \cite{badrinarayanan2015segnet} & \url{https://mi.eng.cam.ac.uk/projects/segnet}\\\hline
Brain lesion synthesis using GANs & \cite{shin2018medical} & \url{https://github.com/khcs/brain-synthesis-lesion-segmentation} \\\hline
GANCS: Compressed Sensing MRI based on Deep Generative Adversarial Network & \cite{mardani2017deep} & \url{https://github.com/gongenhao/GANCS}\\\hline
Deep MRI Reconstruction & \cite{Schlemper2018} & \url{https://github.com/js3611/Deep-MRI-Reconstruction}\\\hline
Graph Convolutional Networks for brain analysis in populations, combining imaging and non-imaging data& \cite{parisot2017spectral} & \url{https://github.com/parisots/population-gcn}\\ \hline

\end{longtable}
}

{\footnotesize
\centering

\begin{longtable}{|p{2cm}|p{6cm}|p{6cm}|} 
\caption{A short list of medical imaging data sets and repositories} \label{tab:data}\\
%\hline\\
\textbf{Name} & \textbf{Summary} & \textbf{Link}\\\hline
OpenNeuro & An open platform for sharing neuroimaging data under the public domain license. Contains brain images from 168 studies (4,718 participants) with various imaging modalities and acquisition protocols. & \url{https://openneuro.org}\footnote{Data can be downloaded from the AWS S3 Bucket \url{https://registry.opendata.aws/openneuro}.} \\\hline
UK Biobank & Health data from  half a million participants. Contains MRI images from 15.000 participants, aiming to reach 100.000. & \url{http://www.ukbiobank.ac.uk/} \\\hline
TCIA & The cancer imaging archive hosts a large archive of medical images of cancer accessible for public download. Currently contains images from from 14.355 patients across 77 collections. & \url{http://www.cancerimagingarchive.net} \\\hline
ABIDE & The autism brain imaging data exchange. Contains 1114 datasets from 521 individuals with Autism Spectrum Disorder and 593 controls.  & \url{http://fcon_1000.projects.nitrc.org/indi/abide} \\\hline
ADNI & The Alzheimer's disease neuroimaging initiative. Contains image data from almost 2000 participants (controls, early MCI, MCI, late MCI, AD)   & \url{http://adni.loni.usc.edu/}\\\hline

\end{longtable}

}

{\footnotesize
\centering

\begin{longtable}{|p{4cm}|p{4.3cm}|p{5.7cm}|} 
\caption{A short list of medical imaging competitions} \label{tab:competitions} \\
%\hline\\
\textbf{Name} & \textbf{Summary} & \textbf{Link}\\\hline
Grand-Challenges & Grand challenges in biomedical image analysis. Hosts and lists a large number of competitions & \url{https://grand-challenge.org/} \\\hline
RSNA Pneumonia Detection Challenge & Automatically locate lung opacities on chest radiographs & \url{https://www.kaggle.com/c/rsna-pneumonia-detection-challenge}\\\hline
HVSMR 2016 & Segment the blood pool and myocardium from a 3D cardiovascular magnetic resonance image & \url{http://segchd.csail.mit.edu/} \\\hline
ISLES 2018 & Ischemic Stroke Lesion Segmentation 2018. The goal is to segment stroke lesions based on acute CT perfusion data.  & \url{http://www.isles-challenge.org/} \\ \hline
BraTS 2018 &Multimodal Brain Tumor Segmentation. The goal is to segment brain tumors in multimodal MRI scans. & \url{http://www.med.upenn.edu/sbia/brats2018.html} \\\hline
CAMELYON17 &The goal is to develop algorithms for automated detection and classification of breast cancer metastases in whole-slide images of histological lymph node sections. & \url{https://camelyon17.grand-challenge.org/Home}\\\hline
ISIC 2018 &Skin Lesion Analysis Towards Melanoma Detection &\url{https://challenge2018.isic-archive.com/} \\ \hline
Kaggle's 2018 Data Science Bowl&Spot Nuclei. Speed Cures. & \url{https://www.kaggle.com/c/data-science-bowl-2018}\\ \hline
Kaggle's 2017 Data Science Bowl&Turning Machine Intelligence Against Lung Cancer & \url{https://www.kaggle.com/c/data-science-bowl-2017}\\ \hline
Kaggle's 2016 Data Science Bowl &Transforming How We Diagnose Heart Disease & \url{https://www.kaggle.com/c/second-annual-data-science-bowl} \\\hline
MURA & Determine whether a bone X-ray is normal or abnormal & \url{https://stanfordmlgroup.github.io/competitions/mura/}\\ \hline
\end{longtable}

}

\section{Challenges, limitations and future perspectives}
It is clear that deep neural networks are very useful when one is tasked with producing accurate decisions based on complicated data sets. But they come with some significant challenges and limitations that you either have to accept or try to overcome. Some are general: from technical challenges related to the lack of mathematical and theoretical underpinnings of many central deep learning models and techniques, and the resulting difficulty in deciding exactly what it is that makes one model better than another, to societal challenges related to maximization and spread of the technological benefits \cite{marcus2018deep, Lipton2018} and the problems related to the tremendous amounts of hype and excitement\footnote{Lipton: \textit{Machine Learning: The Opportunity and the Opportunists} \url{https://www.technologyreview.com/video/612109}, Jordan: \textit{Artificial Intelligence -- The Revolution Hasn't Happened Yet} \url{https://medium.com/@mijordan3/artificial-intelligence-the-revolution-hasnt-happened-yet-5e1d5812e1e7}}. Others are more domain-specific.

In deep learning for standard computer vision tasks, like object recognition and localization, powerful models and a set of best practices have been developed over the last few years.  The pace of development is still incredibly high, but certain things seem to be settled, at least momentarily. Using the basic building blocks described above, placed according to the ideas behind, say, ResNet and SENet, will easily result in close to state-of-the-art performance on two-dimensional object detection, image classification and segmentation tasks. 

However, the story for deep learning in medical imaging is not quite as settled. One issue is that medical images are often three-dimensional, and three-dimensional convolutional neural networks are as well-developed as their 2D counterparts. One quickly meet challenges associated to  memory and compute consumption when using CNNs with higher-dimensional image data, challenges that researchers are trying various approaches to deal with (treating 3D as stacks of 2Ds, patch- or segment-based training and inference, downscaling, etc). It is clear that the ideas behind state-of-the-art two-dimensional CNNs can be lifted to three dimensions, but also that adding a third spatial dimension results in additional constraints. Other important challenges are related to data, trust, interpretability, workflow integration, and regulations, as discussed below.

\subsection{Data} This is a crucially important obstacle for deep neural networks, especially in medical data analysis. When deploying deep neural networks, or any other machine learning model, one is instantly faced with challenges related to data access, privacy issues, data protection, and more. 

As privacy and data protection is often a requirement when dealing with medical data, new techniques for training models without exposing the underlying training data to the user of the model are necessary. It is not enough to merely restrict access to the training set used to construct the model, as it is easy to use the model itself to discover details about the training set \cite{zhang2016understanding}. Even hiding the model and only exposing a prediction interface would still leave it open to attack, for example in the form of model-inversion \cite{fredrikson2015model} and membership attacks \cite{shokri2017membership}. Most current work on deep learning for medical data analysis use either open, anonymized data sets (as those in Table \ref{tab:data}), or locally obtained anonymized research data, making these issues less relevant. However, the general deep learning community are focusing a lot of attention on the issue of privacy, and new techniques and frameworks for \textit{federated learning} \cite{pmlr-v54-mcmahan17a}\footnote{See for example \url{https://ai.googleblog.com/2017/04/federated-learning-collaborative.html}}, \textit{split learning} \cite{gupta2018distributed, vepakomma2018split} and \textit{differential privacy} \cite{papernot2016semi, papernot2018scalable, brendan2018learning} are rapidly improving. See \cite{vepakomma2018no} for a recent survey. There are a few examples of these ideas entering the medical machine learning community, as in \cite{Chang2018} where the distribution of deep learning models among several medical institutions was investigated, but then without considering the above privacy issues. As machine learning systems in medicine grows to larger scales, perhaps even including computations and learning on the ``edge", federated learning and differential privacy will likely become the focus of much research in our community. 

If you are able to surmount these obstacles, you will be confronted with deep neural networks' insatiable appetite for training data. These are very inefficient models, requiring large number of training samples before they can produce anything remotely useful, and labeled training data is typically both expensive and difficult to produce. In addition, the training data has to be representative of the data the network will meet in the future. If the training samples are from a data distribution that is very different from the one met in the real world, then the network's generalization performance will be lower than expected. See \cite{zech2018variable} for a recent exploration of this issue. Considering the large difference between the high-quality images one typically work with when doing research and the messiness of the real, clinical world, this can be a major obstacle when putting deep learning systems into production.

Luckily there are ways to alleviate these problems somewhat. A widely used technique is \textit{transfer learning}, also called fine-tuning or pre-training: first you train a network to perform a task where there is an abundance of data, and then you copy weights from this network to a network designed for the task at hand. For two-dimensional images one will almost always use a network that has been pre-trained on the ImageNet data set. The basic features in the earlier layers of the neural network found from this data set typically retain their usefulness in any other image-related task (or are at least form a better starting point than random initialization of the weights, which is the alternative). Starting from weights tuned on a larger training data set can also make the network more robust. Focusing the weight updates during training on later layers requires less data than having to do significant updates throughout the entire network. One can also do inter-organ transfer learning in 3D, an idea we have used for kidney segmentation, where pre-training a network to do brain segmentation decreased the number of annotated kidneys needed to achieve good segmentation performance \cite{lundervold2017fss}. The idea of pre-training networks is not restricted to images. Pre-training entire models has recently been demonstrated to greatly impact the performance of natural language processing systems \cite{peters2018deep,howard2018universal,radford2018improving}.

Another widely used technique is \textit{augmenting} the training data set by applying various transformations that preserves the labels, as in rotations, scalings and intensity shifts of images, or more advanced data augmentation techniques like anatomically sound deformations, or other data set specific operations (for example in our work on kidney segmentation from DCE-MRI, where we used image registration to propagate labels through a time course of images \cite{lundervold2018feo}). \textit{Data synthesis}, as in \cite{shin2018medical}, is another interesting approach. 

In short, as expert annotators are expensive, or simply not available, spending large computational resources to expand your labeled training data set, e.g. indirectly through transfer learning or directly through data augmentation, is typically worthwhile. But whatever you do, the way current deep neural networks are constructed and trained results in significant data size requirements. There are new ways of constructing more data-efficient deep neural networks on the horizon, for example by encoding more domain-specific elements in the neural network structure as in the \textit{capsule} systems of \cite{hinton2011transforming, sabour2017dynamic}, which adds \textit{viewpoint invariance}. It is also possible to add \textit{attention mechanisms} to neural networks \cite{mnih2014recurrent, xu2015show}, enabling them to focus their resources on the most informative components of each layer input. 

However, the networks that are most frequently used, and with the best raw performance, remain the data-hungry standard deep neural networks.

\subsection{Interpretability, trust and safety} As deep neural networks relies on complicated interconnected hierarchical representations of the training data to produce its predictions, interpreting these predictions becomes very difficult. This is the ``black box" problem of deep neural networks \cite{castelvecchi2016can}. They are capable of producing extremely accurate predictions, but how can you trust predictions based on features you cannot understand? Considerable effort goes into developing new ways to deal with this problem, including DARPA launching a whole program ``\textit{Explainable AI}"\footnote{\url{https://www.darpa.mil/program/explainable-artificial-intelligence}} dedicated to this issue, and lots of research going into enhancing interpretability \cite{olah2018building, montavon2017methods}, and finding new ways to measure sensitivity and visualize features \cite{zeiler2014visualizing, yosinski2015understanding, olah2017feature, hohman2018visual, bach2015pixel}.

Another way to increase their trustworthiness is to make them produce robust uncertainty estimates in addition to predictions. The field of \textit{Bayesian Deep Learning} aims to combine deep learning and Bayesian approaches to uncertainty. The ideas date back to the early 90s \cite{neal1995bayesian, mackay1992practical, dayan1995helmholtz}, but the field has recently seen renewed interest from the machine learning community at large, as new ways of computing uncertainty estimates from state of the art deep learning models have been developed \cite{murray2018,gal2016uncertainty, li2017dropout}. In addition to producing valuable measures that function as uncertainty measures \cite{leibig2017leveraging, kendall2015bayesian, wickstrom2018uncertainty}, these techniques can also lessen deep neural networks susceptibility to \textit{adversarial attacks} \cite{li2017dropout, feinman2017detecting}.

\subsection{Workflow integration, regulations} Another stumbling block for successful incorporation of deep learning methods is \textit{workflow integration}. It is possible to end up developing clever machine learning system for clinical use that turn out to be practically useless for actual clinicians. Attempting to augment already established procedures necessitates knowledge of the entire workflow. Involving the end-user in the process of creating and evaluating systems can make this a little less of an issue, and can also increase the end users' trust in the systems\footnote{The approach we have taken at our MMIV center \url{https://mmiv.no}, located inside the Department of Radiology}, as you can establish a feedback loop during the development process. But still, even if there is interest on the ``ground floor" and one is able to get prototype systems into the hands of clinicians, there are many higher-ups to convince and regulatory, ethical and legal hurdles to overcome. 

\subsection{Perspectives and future expectations}

Deep learning in medical data analysis is here to stay. Even though there are many challenges associated to the introduction of deep learning in clinical settings, the methods produce results that are too valuable to discard. This is illustrated by the tremendous amounts of high-impact publications in top-journals dealing with deep learning in medical imaging (for example \cite{Ganapathy2018, Kermany2018,46425, de2018clinically, Hinton2018, Liu2018, Rieger2018, Chen2018, Zhu2018, Wasserthal2018, Yoo2018}, all published in 2018). As machine learning researchers and practitioners gain more experience, it will become easier to classify problems according to what solution approach is the most reasonable: (i) best approached using deep learning techniques end-to-end, (ii) best tackled by a combination of deep learning with other techniques, or (iii) no deep learning component at all.

Beyond the application of machine learning in medical imaging, we believe that the attention in the medical community can also be leveraged to strengthen the general computational mindset among medical researchers and practitioners, mainstreaming the field of \textit{computational medicine}\footnote{In-line with the ideas of the \textit{convergence} of disciplines and the ``future of health", as described in \cite{Sharp2017}}. Once there are enough high-impact software-systems based on mathematics, computer science, physics and engineering entering the daily workflow in the clinic, the acceptance for other such systems will likely grow. The access to bio-sensors and (edge) computing on \textit{wearable devices} for monitoring disease or lifestyle, plus an ecosystem of machine learning and other computational medicine-based technologies, will then likely facilitate the transition to a \textit{new medical paradigm} that is {\bf p}redictive, {\bf p}reventive, {\bf p}ersonalized, and {\bf p}articipatory - P4 medicine \cite{hood2012personal}\footnote{\url{http://p4mi.org}}.

\section*{Acknowledgements}
We thank Renate Gr\"{u}ner for useful discussions. The anonymous reviewers gave us excellent constructive feedback that led to several improvements throughout the article. Our work was financially supported by the Bergen Research Foundation through the project ``Computational medical imaging and machine learning -- methods, infrastructure and applications".

\vspace{4mm}

\section*{References}

\vspace{-2mm}

{\footnotesize

\begin{multicols}{2}

\bibliographystyle{model1-num-names}

\bibliography{refs/lundervold_zmp,alex_zmp,lundervold_zmp_rev}
\end{multicols}

}

\end{document}